\documentclass{ecai}
\usepackage{times}
\usepackage{graphicx}
\usepackage{latexsym}
\usepackage{subcaption}
\usepackage{hyperref}
\usepackage{xcolor}



\begin{document}

\title{Human Activity Recognition from Wearable Sensor Data Using Self-Attention}
\author{Saif Mahmud \footnotemark[1] \and M Tanjid Hasan Tonmoy \footnotemark[1] 
\and Kishor Kumar Bhaumik \footnotemark[2] \and \\A K M Mahbubur Rahman \footnotemark[2] \and M Ashraful Amin \footnotemark[2]
\and Mohammad Shoyaib \footnotemark[1] \and \\Muhammad Asif Hossain Khan \institute{University of Dhaka, Bangladesh, email: \{2015-116-815, 2015-116-770\}@student.cse.du.ac.bd , \{shoyaib, asif\}@du.ac.bd; The first two authors have equal contributions.} \and Amin Ahsan Ali \institute{Independent University Bangladesh, email: 
\{1621366, akmmrahman, aminmdashraful,aminali\}@iub.edu.bd}
}

\maketitle
\bibliographystyle{ecai}

\begin{abstract}
Human Activity Recognition from body-worn sensor data poses an inherent challenge in capturing spatial and temporal dependencies of time-series signals. In this regard, the existing recurrent or convolutional or their hybrid models for activity recognition struggle to capture spatio-temporal context from the feature space of sensor reading sequence. To address this complex problem, we propose a self-attention based neural network model that foregoes recurrent architectures and utilizes different types of attention mechanisms to generate higher dimensional feature representation used for classification. We performed extensive experiments on four popular publicly available HAR datasets: PAMAP2, Opportunity, Skoda and USC-HAD. Our model achieve significant performance improvement over recent state-of-the-art models in both benchmark test subjects and Leave-one-subject-out evaluation. %In particular, our proposed model has significant improvement of macro F1-score on PAMAP2 and USC-HAD datasets over other state-of-the-art deep learning models. 
% \textcolor{green}{F1-score for our proposed model on PAMAP2 is $0.95$ (sample-wise) and $0.96$ (window-wise) whereas Deep Convolutional long Short Term Memory (DeepConvLSTM),  Convolutional AutoEncoder (ConvAE), and Deep Convolutional LSTM with Attention  have F1-score  well below $0.90$ for both sample-wise and window-wise experiments.  Only LSTM with Continuous Attention  got the F1-score close to 0.90. Similar  performance improvement has also been observed  for USC-HAD dataset.} 
%{In addition, the experimental results also suggest that the proposed model is able to address the issues of varied duration of complex activity as well as diversity in subjects.}
We also observe that the sensor attention maps produced by our model is able capture the importance of the modality and placement of the sensors in predicting the different activity classes.

\end{abstract}

\section{Introduction}
%1st paragraph: What is automatic HAR from sensor modalities? Why automatic HAR necessary?
Human Activity Recognition (HAR) has drawn extensive attention in various areas of mobile health and context-aware computing such as recognition of Nurse care activities \cite{haque2019nurse}, assessment of the quality of physical activities or exercises performed by rehabilitation patients or athletes \cite{panwar}. HAR is defined as the automated classification of the activities of specific subjects wearing heterogeneous sensors placed at different body locations. In other words, HAR takes the readings from different body-worn sensors as input and afterward, it segments and classifies the time-series sensor signal in accordance with the extracted features. Currently, the task of assessment of quality of physical activities or exercises performed by patients is usually performed by an expert physiotherapist. A HAR system can be used to perform this assessment in real-time and assist the healthcare professionals.

%However, developing such a system is challenging due to the within and between subject variations in performing activities. The variations can be considered as the differences both in the spatial and temporal dimension of the feature space, across subjects and even when the same subject performs the same task at different times. 
%Multi-modal sensor-based HAR relies upon processing the signal sequence of the pipelined Activity Recognition Chain (ARC) and consequently mapping the readings of specific time-step to the entities of finite activity class label set. Hence, HAR can be described as a sequence of signal processing, pattern recognition, and machine learning techniques. Multimodal sensor-based HAR is very critical in the context-aware applications in real life such as physical activity logging, cross-device user identification, and health monitoring.

%2 nd paragraph and 3 rd paragraph: What are the current trends towards automatic HAR? Statistical,Convolutional, attention based. Cite 1 or 2 papers for statistical work(concise summary from related works), 3 4 papers for conv work (concise summary from related works), 3-4 papers from attention based models (summary)
%Researches in the last decades include statistical machine learning techniques for automatic HAR.
Although HAR is the core area of wearable and ubiquitous computing, it remains one of the most challenging ones. This is due to large number of sensor modalities, noisy signals, variation in the spatial and temporal dimension of the feature space across subjects and even when the same subject performs the same task at different times, and so on. Researchers from last decades introduced a number of hand-crafted signal processing equations to derive statistical features out of the time-series sensors data. Examples of statistical features are mean, variance, and Fast Fourier Transform coefficients. Then, they used several supervised classification techniques: Support Vector Machines, Decision Trees, Bayesian classifiers to classify the activities \cite{statistical2004activity,plotz2011stfeature}. Later, deep learning based techniques enabled the learning of feature representations for classification tasks without involving domain-specific knowledge. A number of researches have been performed with various architectures of Convolutional Neural Network (CNN) \cite{ha2015multiconv, pourbabaee2017deepconv}. 
Simple physical activities (e.g., walking or cycling) and postures (e.g., sitting or standing) are automatically recognized with good performance with the above-mentioned techniques. However, recognition of many complex activities (stair-up/down, running/jogging, watching TV, ironing) remain challenging. Moreover, sensor displacement and other sources of noise make the HAR more error-prone. 

Recently, hybrid deep learning model consisting of CNN and Recurrent Neural Network (RNN) \cite{Yao:2017:DUD:3038912.3052577, convLSTM:HAR} has achieved better performance with a considerable margin than the conventional CNN model for complex activities as these models consider the activity recognition as a sequence labeling problem. The convolution and recurrent layers (Gated Recurrent Unit or GRU \cite{Yao:2017:DUD:3038912.3052577} and Long Short-Term Memory \cite{convLSTM:HAR}) together capture the temporal characteristics and relationship among the different sensor modalities. More recently, \cite{murahari2018attention} explores attention module on top of the recurrent layers to improve the performance of the models. In \cite{Zeng:2018:UIR:3267242.3267286, ijcai19:HAR}, the authors explore temporal attention on top of recurrent layers and attention on sensor modality and show that the models improve the performance of HAR on some benchmark human activity datasets. 

%in the task of end-to-end feature extraction for synthetic feature crafting. However, HAR incorporated statistical heuristics \cite{plotz2011feature} to represent the features of the activity recognition chain.

%4 th Paragraph: Summary of limitations in above works. Introduce transformer architecture for NLP 0concisely, then our intention to drag the HAR to transformer architecture because HAR samples, windows, and sessions are analogous to character, word, sentence. Small example might help Figure: block diagram of the proposed architecture
The recurrent neural network based encoder-decoder architectures developed for natural language processing tasks such as Neural Machine Translation (NMT) \cite{sutskever2014sequence} are unable to capture the context from all possible transformed feature combinations. To address this limitation, several research works proposed attention-based mechanism \cite{Bahdanau2015NeuralMT} for NMT where varying attention is given to different words of a sentence. However, recurrent networks are constrained by their sequential operations. These limitations have led the researchers towards innovating Transformer architecture \cite{NIPS2017_7181} for NMT.

Transformer leverages self-attention \cite{Lin-self-attn} which enables the model to capture context within the sequence. Transformer avoids the sequential processing involved in recurrent architectures and depends solely on self-attention and positional encoding technique. Transformer also includes multi-headed architecture in order to capture self-attention from different perspectives. Thus, transformer architecture plays important role in capturing context through weight distribution in the temporal dimension and in computing attention in natural language modeling. With the similar idea, we adopt self-attention architecture from NMT task for HAR and propose a model incorporating self-attention with sensor and temporal attention. 

%5 th paragraph: How our goal/contribution/intention can solve HAR problem more efficiently compared to above techniques. How and why fully attention based technique can facilitate to overcome the existing challenges. Describe the proposed architecture here.

%transformer
 In this paper, we propose that sensor's data samples are equivalent to words and windows (time window) are analogous to sentence. Hence, the objective of this paper is to build an attention based end-to-end system where attention is utilized in different ways to create effective feature representation of sensor data. To do that, we introduce the first attention layer on the raw input. Secondly, we adopted self-attention and positional encoding from the transformer architecture \cite{NIPS2017_7181} for HAR to capture spatio-temporal dependencies of sensor signals and their modalities. After a number of self attention blocks, we add another layer of attention that facilitates learning of global attention from the context. Finally, a fully connected layer is placed to classify the activity.

%6 th Paragraph: specifically describe the Summary of works that we have done in this paper. What databases and experiments we used to show the efficacy of the proposed method. How the comparison looks like. Concise summary of comparison and the final conclusive result. Hence our contributions are: Enlist contribution with numbering

In this work, we have experimented with four benchmark human activity recognition datasets: PAMAP2 \cite{Reiss:2012:INB:2357489.2358027}, USC-HAD \cite{usc-had_paper}, Opportunity \cite{opportunity-dataset}, and SKODA \cite{skoda-paper} and compared our results with the current state-of-the-art techniques namely DeepConvLSTM \cite{convLSTM:HAR} and Convolutional Autoencoder (ConvAE) \cite{onFeature_ISWC2019} to demonstrate the effectiveness of the proposed approach. We observe that the proposed model outperforms the DeepConvLSTM and ConvAE for both sample-wise and window-wise experimental setup on benchmark test-cases from the aforementioned datasets. We also perform leave one subject out cross validation experiments to show the superiority of our proposed model for generalization across subjects.
Hence, our contributions are enlisted below:
\begin{enumerate}
 \item We propose a self-attention based non-recurrent neural network architecture for HAR.
 \item We incorporate sensor modality attention and global temporal attention at different layers. The attention layers capture the spatio-temporal context in the sensor signal to construct feature representation for classification.
 \item We compare our model with other state of the art models on four publicly available HAR datasets in terms of both benchmark test sets and leave one subject out tests. In addition, we analyze the impact of various window-size on the proposed and other existing models.
 \item We construct sensor-level attention maps that are intuitively explainable and thus demonstrates interpretability of the modules of the architecture.
\end{enumerate}
%7 th Paragraph: outline of the rest of the paper.
The rest of this paper is organized as follows. Section \ref{rel-work} introduces related works on HAR using deep learning and various attention models. We describe the technical details of our proposed model in Section \ref{proposed-method}. In Section \ref{dataset-description} and \ref{exp-setup}, we describe the datasets and the setup of the experiments, respectively. Comparative results with existing models and other experiments are presented in Section \ref{results}. Finally, we conclude the paper in Section \ref{conclusion}.

\section{Related Works}
\label{rel-work}
%machine learning
Plethora of research has been conducted in the area of human activity recognition since 2000. 
Recently, Haresamudram et al. \cite{onFeature_ISWC2019} presented a comprehensive review of deep learning based feature extraction and recognition models for HAR using sensor data. In the last decade, most of the wearable device-based HAR involves hand-crafted features from domain-specific knowledge in the case of shallow machine learning models. These models depend on statistical features \cite{Figo:2010:PTC:1872968.1872975,plotz2011stfeature} and distribution-based features \cite{kwon2018adding,hammerla2013preserving}. Statistical features are calculated using different statistical characteristics equations \cite{Figo:2010:PTC:1872968.1872975}. There also has been efforts to use LSTM with attention on statistical and geometric features in the context of HAR based on 3D skeleton data \cite{haque2019nurse,zhang2018fusing}. On the other hand, distribution-based feature representations are obtained from signal processing approaches \cite{Huynh:2005:AFA:1107548.1107591} such as wavelet or Fourier transformation.

%deep learning
Recently, end-to-end deep learning based techniques for HAR have gained more popularity among the machine learning communities. As the deep learning based techniques find the most appropriate feature representations for the HAR with supervised fashion, they have eliminated human-intervened feature crafting and data representation tuning through simultaneous representation learning and classifier optimization \cite{onFeature_ISWC2019}. Convolutional neural network \cite{Hammerla:2016:DCR:3060832.3060835} and its combination with recurrent networks (DeepConvLSTM) \cite{convLSTM:HAR} have shown notable performance in capturing spatial-temporal features from the sensor signal data.

%attention mechanism
Furthermore, the utilization of attention mechanism for HAR has been explored in \cite{murahari2018attention, gru_attn_icasert} by combining it with recurrent networks. In particular, the DeepConvLSTM architecture proposed in \cite{convLSTM:HAR} has been augmented with an attention layer in \cite{murahari2018attention}. This layer learns parameters to compute the relative weights for the hidden state outputs of the preceding LSTM layer. Attention layer is used to create context vector using linear combination of past and current hidden states in contrast to \cite{convLSTM:HAR} which uses the last hidden state as context. Integration of continuous temporal and modality attention with LSTM has been proposed in \cite{Zeng:2018:UIR:3267242.3267286}. In the same way, the augmentation of attention in two capacities \cite{ijcai19:HAR} is proposed to compute the relative weight of sensor modality for specific activity window and to encapsulate the temporal context of the salient features of specific sensor signal. This approach, based on attention augmented GRU and ConvNet architecture, makes use of overlapping sliding window of Fast Fourier Transformed spectrogram from sensor signals. This recurrent architecture based attention model \cite{ijcai19:HAR} is referred to as AttnSense and obtains notable performance in temporal context capturing. In this regard, the existing attention models for HAR exhibits notable performance in adapting inter and intra activity class variance with adaptive duration of attention within activity sequence. Zheng et al. \cite{icann} has proposed uniqueness attention based LSTM architecture which captures atomic features in temporal context. However, no architecture has been proposed yet which incorporates self-attention to capture spatial context of the feature sequence along with temporal context capturing.
% Nevertheless, the model has not been able to obtain best performance with respect to F-measure for benchmark human activity dataset despite capturing noise pattern and spatial context of feature representations.

%Attention based HAR models

%\section{State-of-the-art Convolutional and Recurrent Models for HAR}
%\label{sota-HAR}

\section{Proposed Self-attention Model}
\label{proposed-method}
\begin{figure}[t]
	\centering
	\includegraphics[width=0.95\columnwidth]{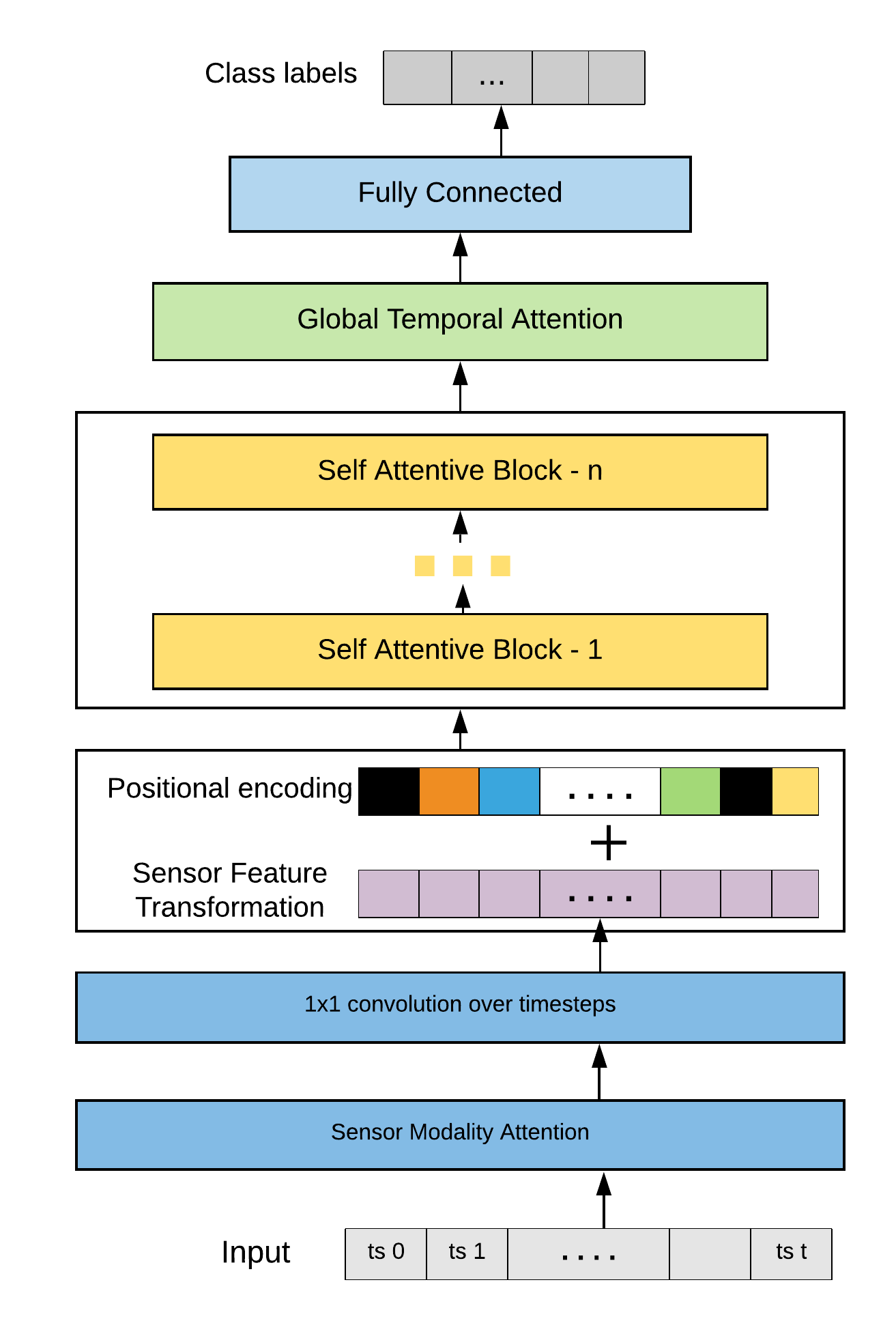}
	\caption{Attention based model incorporating self-attention and global temporal attention}
	\label{fig:model-transf}
\end{figure}

Our objective is to build a self-attention based model without any recurrent architectures. Hence, the proposed model foregoes recurrent networks and utilizes sensor modality attention, self attention blocks, global temporal attention to construct feature representation used for classification as illustrated in Figure \ref{fig:model-transf}. We briefly describe the model architecture below and provide detailed specification in the subsequent section.

\label{proposed:description}
The input to the model is a time-window of sensor values. Firstly, sensor modality attention is applied to the inputs to get a weighted representation of the sensor values according to their attention score. Thus, the learned attention score represents the contribution of each of the sensor modalities in the feature representation used by the subsequent layers. Afterwards, we convert the weighted sensor values to $d$ size vectors using 1-D convolution over single time-steps. Similar to \cite{NIPS2017_7181}, we encode positional information of the samples in the sequence by adding values based on sine and cosine functions to the obtained $d$ size vectors. This enables the model to take the temporal order of samples into account. This representation is scaled by $\sqrt{d}$ and passed to the self attention blocks. Self attention blocks use dot product-based attention score to transform the feature value for each time step. The representation generated from the self attention blocks is used by global temporal attention layer. As shown in Figure~\ref{fig:model-transf}, this layer learns parameters to set varying attention across the temporal dimension to generate the final representation which is used by the final fully connected and softmax layers. We discuss details of sensor modality attention, self attention blocks, global temporal attention modules in the following  subsections.

\subsection{Sensor Modality Attention}
\label{proposed:sensor-modality}
%why and how-intro
To capture the varying levels of contribution from sensors at different modalities for classification, we use the sensor modality attention layer. For example, in order to recognize the activity 'ironing', the sensors placed at the subject's ankle do not provide much meaningful information. Sensor attention layers learn such relationships by using 2-d convolution across time-step and sensor values to capture their dependencies.

%detail
Firstly, the input is reshaped to produce single channel image. Then, $k$ convolutional filters are applied to the input with padding which outputs image with k channels. This is then converted back to a single channel by applying $1\times 1$ convolution. Sensor-wise softmax as defined in (\ref{eqn:row-softmax}) provides the attention score for individual sensors. In addition to providing a weighted version of the input according to their learned importance for the self attention layer, this mechanism allows us to plot feature maps making the model more interpretable. 

\begin{equation}
\label{eqn:row-softmax}
s_{\kappa} ^{(t_i)} = \frac{\exp ({q_\kappa^{(t_i)}})}{\Sigma _\kappa \exp ({q_\kappa^{(t_i)}})}
\end{equation}

$\kappa$ in (\ref{eqn:row-softmax}) indicate individual sensors.

\subsection{Self Attention Block}
\label{proposed:self-attentive}

Each block consists of two sub layers - multi-headed self attention and position-wise feed forward layer. Self attention is used to determine relative weights for each time-step in the sequence by considering its similarity to all the other time-steps. Subsequently, these relative weights are used to transform the representation of each time-step with relevant information from other time-steps according to their importance.
\begin{equation}
\label{eqn:sa}
f_{sa}^{(h_j)}(\mathbf{q,k,v}) = softmax( \frac{\mathbf{q} \cdot \mathbf{k}^T}{\sqrt{d_k}})\mathbf{v}
\end{equation} 

The terms (q, k, v) in (\ref{eqn:sa}) are learned linear transformation of the input to the layer and referred as key, query and value respectively. In this regard, the query can be considered to the transformed vector of a particular time-step that is compared to the key vector of every other time-step using dot product. Afterwards, the dot product value is scaled and softmax normalized which indicates the attention scores. Finally, the attention values are used to get a weighted representation of the value vectors for each of the time-steps. However, the entire operation is implemented as a matrix multiplication operation as indicated in (\ref{eqn:sa}). %should we include eqn for k, q, v?

Moreover, we utilize multi-headed self attention since different attention heads are able to capture distinct aspects of the input signal. In this regard, $h_j$ in (\ref{eqn:sa}) represents output from attention head $j$. For computing the key, query and value used in (\ref{eqn:sa}), each one of the $n$ attention heads use distinct parameters. The outputs from the distinct attention heads are concatenated and converted back to the dimension of single attention head using learned parameter $W_o$ as defined in (\ref{eqn:sa-multi-head}).

\begin{equation}
    \label{eqn:sa-multi-head}
    \mathbf{s}_{mha} = \mathbf{W}_o \cdot \textnormal{concat}(\mathbf{f}_{sa}^{(h_1)}\textsc{, ..... ,}\mathbf{f}_{sa}^{(h_{n-1})}, \mathbf{f}_{sa}^{(h_n)})
\end{equation}

Position-wise feed forward layer is applied independently to each position in a block. In this case, the weights are for each position in a block but different across the blocks.

Each of the sub-layer contains a residual connection and is followed by layer normalization.

\subsection{Global Temporal Attention}
\label{proposed:global-temporal}

We use the representation generated by the self attention blocks for each time-step and learn parameters to rank them according to their respective importance for predicting the corresponding class label for the window. The ranking (attention score) obtained in (\ref{eq2-c-attn}) is used to create a weighted average of the representations of all the time-steps in an activity window which is used as feature vector by the feed forward layers for classification.

\begin{equation}
\label{eq1-c-attn}
\mathbf{g}^{(t_i)} = \tanh{(\mathbf{W}_{ga}\cdot \mathbf{s}^{(t_i)} + \mathbf{b}_{ga})}
\end{equation}

\begin{equation}
\label{eq2-c-attn}
\mathbf{\alpha} ^{(t_i)} = \frac{\exp((\mathbf{g}^{(t_i)})^T \cdot \mathbf{g}_s)}{\sum _t \exp (\mathbf{g}^{(t_i)} \mathbf{g}_s)}
\end{equation}

\begin{equation}
\label{eq3-c-attn}
\mathbf{c}_{i} = \sum _t \mathbf{\alpha} ^{(t_i)} \mathbf{s}^{(t_i)}
\end{equation}

The terms $W_{ga}$ and $b_{ga}$ in (\ref{eq1-c-attn}) refer to parameters learned during training to get a hidden representation from each of the vectors $s^{(t_i)}$ generated from self attention blocks. The parameter $g_s$ in (\ref{eq2-c-attn}) helps to capture temporal context while learning to compute the attention score. A weighted summation according to the relative importance of respective time steps is generated as the feature vector in (\ref{eq3-c-attn}).

For regularization, dropout has been used in the self attention blocks, the fully connected layers, and after the addition of positional encoding.
 
\begin{table*}[ht]
	\begin{center}
		{\caption{Summary of experimental setup for the datasets. Here A = Accelerometer, G = Gyroscope, M = Magnetometer }
		\label{tab:dataset}}
		\centering
		\begin{tabular}{lccccc}
			\hline
			\rule{0pt}{12pt}
			Dataset & \begin{tabular}[c]{@{}c@{}}Number of \\ Activities\end{tabular} & \begin{tabular}[c]{@{}c@{}}Benchmark Test \\ Subject ID\end{tabular} & \begin{tabular}[c]{@{}c@{}}Down-\\ sampling \\ \end{tabular} & \begin{tabular}[c]{@{}c@{}}Sliding Window\\ Overlap\end{tabular} & Sensors Used                                                                   \\ \hline \\ [-6pt]
			
			PAMAP2           & 12                   & 106                                                                       & 1/3                 & 50\%                                                                        & A, G                                                                                   \\   \hline
			Opportunity      & 18                   &   \begin{tabular}[c]{@{}c@{}} 2, 3\\ (Run 4 \& 5)\end{tabular}       & 1                   & 50\%                                                                        & \begin{tabular}[c]{@{}c@{}}A, G, M (upper body)\\ \& sensors in shoes\end{tabular} \\  \hline \\ [-6pt]
			USC-HAD          & 12               & 13, 14                                                                    & 1/3                 & 50\%                                                                        & A, G                                                                                   \\  \hline \\ [-6pt]
			Skoda            & 11                   & 1                                                                         & 1/3                 & 50\%                                                                        & A                                                                                      \\ \hline \\ [-6pt]
		\end{tabular}
	\end{center}
\end{table*}

\section{Dataset Description}
\label{dataset-description}
We use four commonly used benchmark datasets  \cite{onFeature_ISWC2019} to evaluate the performance of our model and to compare it with that of state of the art models. However, we did not use Daphnet Freezing of Gait Dataset \cite{daphnetdataset} as this is particular to specific gait recognition for patients with Parkinson's disease. Below We give brief description of the datasets used in our experiments. 

\textbf{PAMAP2} dataset \cite{Reiss:2012:INB:2357489.2358027} incorporates the hardware setup of 3 Inertial Measurement Units (IMU) placed over the wrist of dominant arm, on chest and at ankle and the data has been sampled at the frequency of 100Hz. The whole data included annotated human activity class of 9 subjects with particular physical description. Majority of the subjects are male with right dominant hand. In fact, PAMAP2 contains only one female subject and one left handed subject with id 102 and 108 respectively. This benchmark dataset contains 18 human activity classes altogether. In our experiments, data from one of the accelerometers ($\pm$16g scale) and gyroscope contained in each IMU have been used.

\textbf{OPPORTUNITY} dataset \cite{opportunity-dataset} includes the annotated data of body-worn sensors and ambient sensors to specify particular human activity. The dataset has been formed incorporating the reading of motion sensors and classified with "modes of locomotion". The sensors have been able to capture 5 high-level human activity classes along with 17 mid-level gesture classes and 13 low-level actions. We focus on the mid-level gestures and remaining are considered null class which comprises more than 75\% of the data making the dataset highly imbalanced in terms of class distribution.

\textbf{USC-HAD}
\label{usc-had}
dataset \cite{usc-had_paper} incorporates six readings from body-worn 3-axis accelerometer and gyroscope sensor through Motion-Node device. The dataset has been created with equally distributed (7 each) 14 male and female subjects with defined physical specification and age. The sampling rate of sensor data is 100 Hz and includes one of 12 activity class labels for each time-step in the dataset.

USC-HAD dataset poses an inherent challenge in feature representation learning and segmentation due to the sensor placement and variation in the activity classes. Here, the single accelerometer and gyroscope reading is obtained from the motion node attached to the right hip of specific subject and thus does not contribute much in the feature space transformation. Moreover, the activity classes involve orientation such as walking forward or left or right and even elevator up or down which are generally not captured only through accelerometer and gyroscope reading.

\textbf{SKODA}
\label{skoda}
dataset \cite{skoda-paper} is a special purpose dataset to track the activity of workers in the manufacturing assembly-line scenario. This dataset incorporates accelerometer reading from 10 different positions on the subject's arms and is labeled with specific activity class including a null class. Following the standard procedure, we use 80\% of the data for training and 10\% for validation and test respectively. 

The benchmark test subjects and the summary of the datasets have been included in Table \ref{tab:dataset}.

\section{Experiment Setup}
\label{exp-setup}
In this section we describe the preprocessing of the datasets, the architecture of the models, evaluation procedure, and performance measures used in our experiments. 
%describe how you preprocess and and what format you feed the data to %models. General processing that are common for all models
\subsection{Preprocessing}
%Describe here how you make windows, what dimension you are representing the sensor modalities. Which dimension you are using for time steps
Since the datasets involved in the experiments have varied sampling rates, alignment of the frequencies through downsampling facilitates reasonable comparison of performance. Similar to the previous works in \cite{onFeature_ISWC2019} and \cite{Zeng:2018:UIR:3267242.3267286}, we down-sampled PAMAP2, USC HAD and Skoda to close 30 Hz to align with the Opportunity dataset.

\textbf{Window based representation: }
%What is the window size. How you make each window tat looks like a 2D image.
The proposed approach utilizes sliding window based feature extraction. Window size is the number of samples that is considered at a time to construct a feature representation used for classification. The activities under consideration are diverse in terms of duration and complexity which makes the choice of window size an important hyper-parameter. Likewise, the choice of how much overlap there should be between the consecutive windows is also an important factor to consider.

\begin{figure}[h!]
    \centering
    \includegraphics[width=.75\columnwidth]{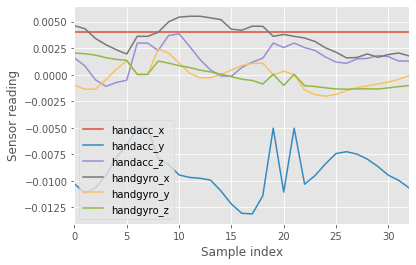}
    \caption{Activity window for walking activity in PAMAP2 dataset where timespan = 1 Sec}
    \label{fig:window_visualization_PAMAP2_walking}
\end{figure}

The activity recognition window is constructed like an image with time-steps and heterogeneous sensor readings as the two dimensions. The activity label is determined through majority voting in the samples constituting the window. The activity window with specific class label has been demonstrated in Figure \ref{fig:window_visualization_PAMAP2_walking}.

%How you slide your window. How much overlaps, What are the overlaps.
The time-series sensor data indicates activities spanning varied time-span and to capture this sequence within the activity recognition window, the sliding of windows is done with a fixed percentage of overlap. The percentage of overlap has been tuned as a hyperparameter and deployed in the training accordingly.

\subsection{Implementation of Existing Architectures}
We perform extensive experiments with the Convolutional Autoencoder (ConvAE) and Deep Convolutional LSTM (DeepConvLSTM) \cite{convLSTM:HAR}. Experiments presented in \cite{onFeature_ISWC2019} show that these models perform well for different benchmark datasets. We also report the experimental results for newly published attention based HAR models from the respective papers. Below we provide detailed description of the ConvAE and DeepConvLSTM models specified in \cite{onFeature_ISWC2019} pertaining to our experiments.

\textbf{Convolutional Autoencoder (ConvAE)} includes an encoder and a decoder part with a bottleneck layer in between. The encoder consists of four convolutional blocks each containing two $3\times3$ convolution layers with the same number of filters followed by batch normalization. Each of the convolutional blocks contain a $2\times2$ max-pooling operation at the end. The output from the last convolutional block is flattened and passed to a fully connected bottleneck layer. The feature vector from the bottleneck layer is used by the decoder to reconstruct the input by inverting the encoding process sequentially. We followed the description of the decoder in \cite{onFeature_ISWC2019} and used up-sampling, convolution and appropriate padding or cropping to match the input dimension during the inversion process. Similar to the encoder, we used the same number of $3\times3$ filters in the four blocks of the decoder. We used relu activation throughout the model and hyperbolic tangent activation for the output. The feature representation from the bottleneck layer is used by Multi-layer Perceptron (MLP) \cite{onFeature_ISWC2019} to classify the activity label for the respective input window. The dimension of the bottleneck layer has been set to 500, 1000, 1500 and 2000 for PAMAP2, Opportunity, Skoda and USC HAD respectively since the best results for the respective datasets have been reported at that particular dimension in \cite{onFeature_ISWC2019}.

\textbf{DeepConvLSTM} has four successive convolution layers and two layers of LSTMs. Each convolutional layer has 64 filters with size of $5\times1$. The $5\times1$ filter is used to perform convolution along the time-steps. Using the $5\times1$ filter, multiple sensor information are kept separate. The output from the first convolution layer is fed to the 2nd convolution layer and so on. Then the output of the final convolution layer is applied to a two-layer LSTM, each with 128 hidden units. The final output vector is connected to a fully connected layer. After performing the softmax operation on fully connected layer output, activity class probability is available in the final output of the model. We use a dropout with probability of 0.5 in the fully connected layer.

Detailed description of the architectures of the attention based models namely, DeepConvLSTM with Attention, LSTM with continuous Attention and AttnSense can be found in \cite{murahari2018attention, Zeng:2018:UIR:3267242.3267286, ijcai19:HAR} respectively.

\begin{table*}[ht]
	\begin{center}
		\caption{Macro F1-score for benchmark test subject}
		\label{benchmark-result}
		\centering
		\begin{tabular}{lccccccccc}
			\hline
			\rule{0pt}{12pt}
			&\multicolumn{9}{c}{Architecture} \\ [-2pt]
			 & \multicolumn{2}{c}{Proposed Model}                                                                                      & \multicolumn{2}{c}{DeepConvLSTM}                                                                                        & \multicolumn{2}{c}{ConvAE}                                                                                              & \begin{tabular}[c]{@{}c@{}}DeepConvLSTM\\ + Attention\end{tabular} & \begin{tabular}[c]{@{}c@{}}LSTM\\ + Continuous\\ Attention\end{tabular} & AttnSense   \\ \cline{2-10}
			 \rule{0pt}{12pt}
			\quad Dataset      & $\Diamond$&$\Box$ & $\Diamond$&$\Box$ & $\Diamond$&$\Box$ & $\Diamond$                                                    & $\Diamond$                                                             & $\Diamond$ \\ \hline \\ [-6pt]
			\quad PAMAP2                & 0.95                                                            & 0.96                                                            & 0.71                                                            & 0.70                                                            & 0.52                                                            & 0.80                                                            & 0.88                                                                    & 0.90                                                                             & 0.89                 \\ 
			\quad Opportunity           & 0.61                                                            & 0.67                                                            & 0.66                                                            & 0.58                                                            & 0.60                                                            & 0.60                                                            & 0.71                                                                    & -                                                                                & -                    \\ 
			\quad USC-HAD               & 0.50                                                            & 0.55                                                            & 0.42                                                            & 0.38                                                            & 0.42                                                            & 0.46                                                            & -                                                                       & -                                                                                & -                    \\ 
			\quad Skoda                 & 0.93                                                            & 0.97                                                            & 0.96                                                            & 0.88                                                            & 0.82                                                            & 0.79                                                            & 0.91                                                                    & 0.94                                                                             & 0.93                 \\ \hline
			\\[-6pt]
			\multicolumn{8}{l}{$\Diamond$ Sample-wise\ \
				$\Box$ Window-wise}
		\end{tabular}
	\end{center}
\end{table*}

%How you pad the samples if some samples in the last window are empty
\subsection{Training and Test Procedures}
%How you output window based
For the segmentation of activity data, we have the choice of predicting class label each individual sample in a sequence or for a fixed-length time window. However, we need to analyze a sequence window of some length in both cases. 
\\
\textbf{Sample-wise:} During training, we slide the window by one time-step forward and provide the ground truth label for each time step. Then we slide the window right by one time step. During the testing, we follow the same technique. We take the output label of the window and set this output to corresponding last time-step of the window. Hence, we obtain sample-wise output during test.
\\
\textbf{Window-wise:}
We create the window with predefined window size and continue to slide the window with $50\%$ overlap. During the training, we will assign the most frequent activity in this window as the ground truth label of that window. In testing, the model produces one output label for each window. For test, no overlap is used. In the case where a window contains samples with a different label, we pad the window by repeating the last few samples and the next window starts from the differently labeled sample.
\\
\textbf{Training and Hyperparameters:} For the proposed model, we set the number of self attention blocks to $2$ for all of the datasets except USC-Had where $3$ blocks are used. For construction of fixed size input for self attention as described in Section \ref{proposed:description}, $d$ was set to 128. Similar to \cite{NIPS2017_7181}, the number of units in position-wise feed forward layer was set to $4$ times $d$. We used Adam optimizer with the default parameters discussed in \cite{Kingma2015AdamAM} and learning rate $0.001$ for training of the models.
%How you want to compare
%How you design your experiment goals. What to see
%\subsection{Performance Evaluation}

% \begin{table*}[h!]
% \caption{Average Macro F1-score for leave one subject out experiment}
% \label{avg-leave-one-out-result}
% \centering
% \begin{tabular}{|c|c|c|c|c|c|c|}
% \hline
% \textbf{Architecture} & \multicolumn{2}{c|}{\textbf{Proposed Model}} & \multicolumn{2}{c|}{\textbf{DeepConvLSTM}} & \multicolumn{2}{c|}{\textbf{ConvAE}} \\ \hline
% \textbf{Dataset}      & \textbf{Sample-Wise}  & \textbf{Window-Wise} & \textbf{Sample-Wise} & \textbf{Window-Wise} & \textbf{Sample-Wise}  & \textbf{Window-Wise}  \\ \hline
% PAMAP2                &   0.92               & 0.96                 & 0.61                     &0.52                       &0.47                      & 0.48                      \\ \hline
% Opportunity           & 0.39                   & 0.42                 & 0.44                     & 0.41                      & 0.41                 &0.42                       \\ \hline
% USC-HAD               & 0.60                  & 0.67                 &0.59                      &0.50                      & 0.58                      &   0.63                    \\ \hline
% \end{tabular}
% \end{table*}

\begin{table}
	\begin{center}
	{\caption{Average Macro F1-score for leave one subject out experiment}
	\label{avg-leave-one-out-result}}
	\centering
	\begin{tabular}{lcccccc}
		\hline
		\rule{0pt}{12pt}
		& \multicolumn{6}{c}{Architecture} \\
		& \multicolumn{2}{c}{Proposed Model} & \multicolumn{2}{c}{DeepConvLSTM} & \multicolumn{2}{c}{ConvAE} \\ \cline{2-7}
		\rule{0pt}{12pt}
		\quad Dataset      & $\Diamond$&$\Box$ & $\Diamond$&$\Box$ &$\Diamond$&$\Box$  \\ \hline
		\\ [-6pt]
		\quad PAMAP2                &   0.92               & 0.96                 & 0.61                     &0.52                       &0.47                      & 0.48                      \\ 
		\quad Opportunity           & 0.39                   & 0.42                 & 0.44                     & 0.41                      & 0.41                 &0.42                       \\
		\quad USC-HAD               & 0.60                  & 0.67                 &0.59                      &0.50                      & 0.58                      &   0.63                    \\ \hline
		\\[-6pt]
		\multicolumn{7}{l}{$\Diamond$ Sample-wise\ \
			$\Box$ Window-wise}
	\end{tabular}
\end{center}
\end{table}

\subsection{Evaluation Metric}
We use macro average F1-score as metric to compare the performance of the proposed approach with other methods. In this regard, we calculate F1-score for each class according to (\ref{eqn:f1score}) as follows:
\begin{equation}
\label{eqn:f1score}
\textnormal{Macro F1-Score} = \frac{1}{|C|} *\sum_{i = 1}^{C} \frac{2 * Precision_i * Recall_i}{Precision_i + Recall_i}
\end{equation} 

Here, $ |C| $ in \ref{eqn:f1score} indicates number of classes and F1-score for each class is given the same weight irrespective of their number of instances and $ i = 1, ..., C$ is considered as the set of classes considered for experiment.

\section{Results}
\label{results}
In this section, we present the results of the experiments as described in the previous section. First, we discuss the performance on the benchmark test subjects for the datasets. Then, we describe the Leave One Subject Out Cross Validation (LOSO-CV) results for all subjects. We conclude this section with discussion of the results from window size variation experiments and the attention maps that we generate from the sensor attention layer. 

\textbf{Performance on benchmark test subject:} Table \ref{benchmark-result} shows the performance comparison between the proposed model and the existing models for the benchmark test subjects  described in Table \ref{tab:dataset}.

With regards to the sample-wise classification scores, our proposed model achieves significant improvement over DeepConvLSTM and ConvAE for PAMAP2 and USC-HAD. However, our model obtains slightly lower sample-wise score for Opportunity and Skoda with DeepConvLSTM. 
Specifically, F1 macro has been decreased to $0.61$ from $0.66$ (DeepConvLSTM) for Opportunity dataset. Since the proposed model is fundamentally designed for window based output, we notice the significant improvement while we perform window-based tests for Opportunity.

%Mean F1-macro score across all datasets with proposed model is $0.75$ whereas the same scores for DeepConvLSTM and ConvAE are $0.69$ and $0.59$. Therefore, the mean F1-macro score for the proposed model has been improved by $8\%$ and $25\%$ from DeepConvLSTM and ConvAE, respectively.

However, we can observe more obvious and significant improvement in terms of the window-wise scores for the proposed model. In particular, the window-wise macro F1-score has been improved to $0.67$ from $0.58$ (DeepConvLSTM) and $0.60$ (ConvAE) for the Opportunity dataset. Thus, it can be noted that our model works more accurately ($0.67$) on the Opportunity (datasets containing complex activities) compared to other methods. In terms of the window-based scores, our model also outperforms other models. Therefore, we can conclude that the proposed model can better capture the spatio-temporal characteristics of sensor-data more effectively than the DeepConvLSTM and ConvAE.
% The mean macro F1- score for the proposed model are improved by $24.4\%$ (DeepConvLSTM) and $19.2\%$ (ConvAE).

As discussed in Section~\ref{usc-had}, USC-HAD is a particularly challenging dataset due to the sensor setup. However, our model performs better ($0.50$ sample-wise and $0.55$ window-wise) than the other models (DeepCpnvLSTM: $0.42$ \& $0.38$; ConvAE: $0.42$ \& $0.46$).

\begin{figure*}[!t]
    \centering
    \includegraphics[width=0.96\textwidth]{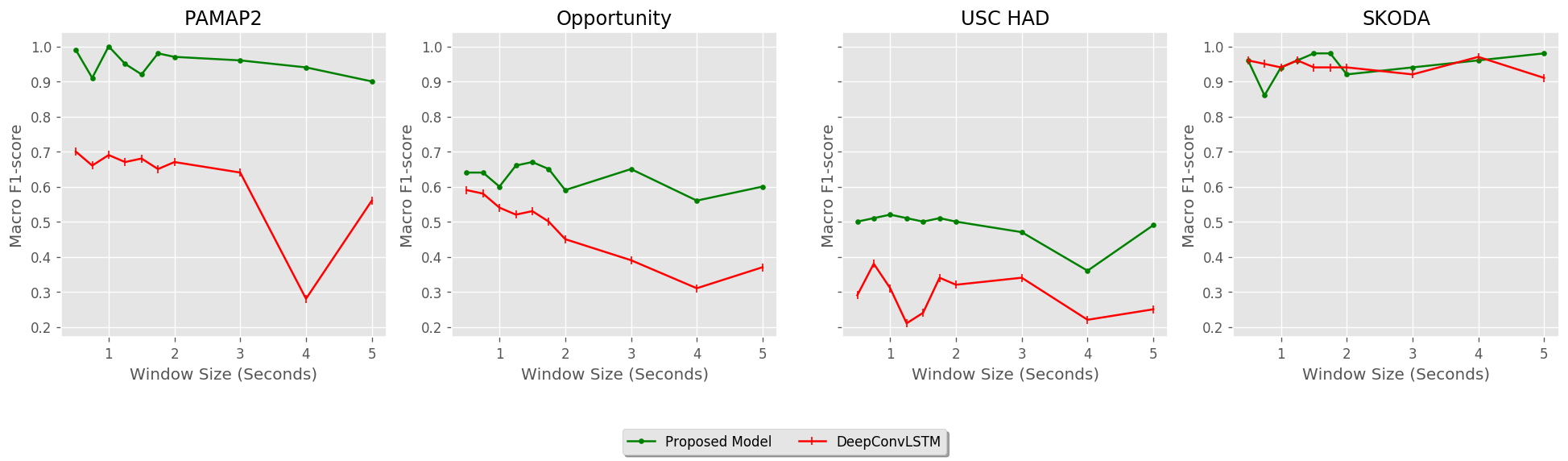}
    \caption{Performance measure against different window sizes}
    \label{fig:window_var}
\end{figure*}

It is evident from the data in Table \ref{benchmark-result} that the proposed model also performs better than other attention based models: DeepConvLSTM + Attention, LSTM + Continuous Attention~\cite{Zeng:2018:UIR:3267242.3267286} and AttnSense~\cite{ijcai19:HAR} for PAMAP2 benchmark test set. We only compare the sample-wise results as the aforementioned models published sample-wise ones only. The macro F1-score for our model for PAMAP2 is higher than the corresponding scores for the other attention-based models e.g DeepConvLSTM + Attention, LSTM + Continuous Attention, and AttnSense. For Skoda, our model also outperformed ($0.93$) DeepConvLSTM + Attention ($0.91$) and equally performed with AttnSense($0.93$). However, our model consistently outperformed the other attention-based models in terms of window-wise test scores on all the datasets considered except Opportunity.

\begin{figure*}[h!]
	\centering
        \begin{subfigure}[]{0.6\columnwidth}
        \centering
        \includegraphics[width=\linewidth]{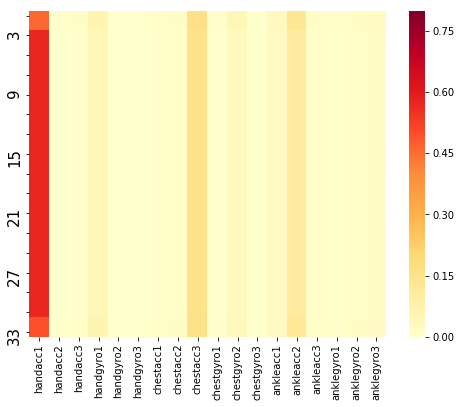}  
        \caption{Ironing}
        \label{fig:sub-first}
        \end{subfigure}
        \begin{subfigure}[]{0.6\columnwidth}
        \includegraphics[width=\linewidth]{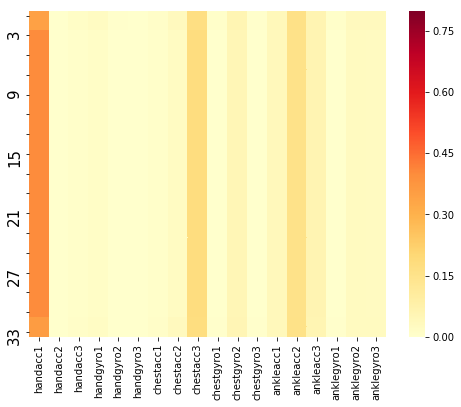}  
        \caption{Ascending stairs}
        \label{fig:sub-second}
        \end{subfigure}
        \begin{subfigure}[]{0.6\columnwidth}
        \includegraphics[width=\linewidth]{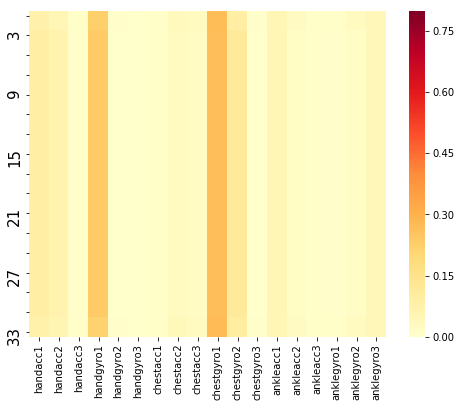}  
        \caption{Lying}
        \label{fig:sub-third}
        \end{subfigure}
        \begin{subfigure}[]{0.6\columnwidth}
        \includegraphics[width=\linewidth]{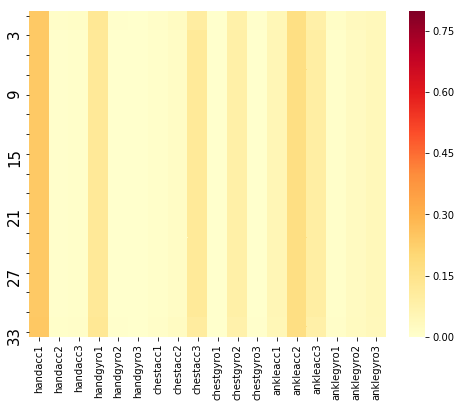}  
        \caption{Cycling}
        \label{fig:sub-fourth}
        \end{subfigure}
        \begin{subfigure}[]{0.6\columnwidth}
        \includegraphics[width=\linewidth]{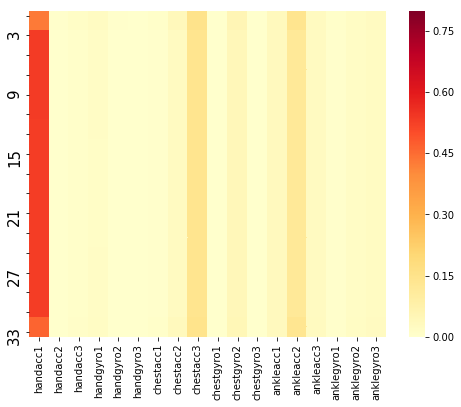} 
        \caption{Vacuum cleaning}
        \label{fig:sub-fifth}
        \end{subfigure}
        \begin{subfigure}[]{0.6\columnwidth}
        \includegraphics[width=\linewidth]{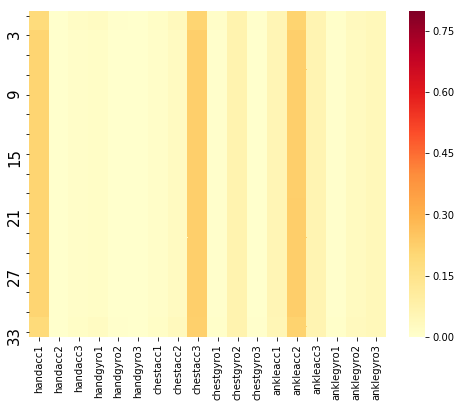} 
        \caption{Walking}
        \label{fig:sub-sixth}
        \end{subfigure}

	\caption{Attention weights on different sensor modality based on predicted class label in PAMAP2 dataset (e.g. Ironing involves higher attention weight for hand accelerometer, moderate attention for chest accelerometer and relatively low weights for other sensor placements)}
	\label{fig:featuremap}
\end{figure*}

\textbf{Performance on LOSO-CV:} In order to demonstrate the robustness of the proposed model in terms of sensitivity to specific test subjects, we conduct LOSO-CV experiments for each dataset containing activities of multiple subjects. In this regard, we hold the data from one of the subjects out during training and use that data for evaluating the model. This process is repeated for each subject for the particular dataset and the average score is reported. Note that, Skoda contains activities performed by only one subject and is excluded from these experiments.

As can be seen from Table \ref{avg-leave-one-out-result}, the macro F1-scores for the proposed method are consistently higher for both sample-wise and window-wise tests compared to the corresponding scores for DeepConvLSTM and ConvAE for LOSO-CV experiments. Thus, the results indicate that the proposed method is capable of modeling the inter-subject variability better. In other words, our model has more generalization capability than the others.

Specifically, LOSO-CV experiments with the PAMAP2 dataset shows that the proposed model significantly outperforms the other models under comparison for subject 102 (female subject). In this regard, the proposed model obtains F1-scores $0.93$ (sample-wise) and $0.98$ (window-wise) respectively. On the other hand, DeepConvLSTM achieves F1-scores of $0.47$ (sample-wise) and $0.33$ (window-wise) in the LOSO-CV experiment involving this subject. For ConvAE, the corresponding score is $0.35$ in both cases.

Moreover, for subject 108 (male, left-handed), our model achieves F1 scores of $0.79$ (sample-wise) and $0.88$ (window-wise) whereas DeepConvLSTM gets $0.27$ and $0.28$, respectively. ConvAE performs slightly better than DeepConvLSTM, the scores for ConvAE are $0.40$ (sample-wise) and $0.47$ (window-wise). 

\textbf{Performance of proposed model on window sizes:}
As different activities have different repetitive periods, we conducted experiments to analyze the impact of window-size variations on the proposed model's performance. In this regard, we train different models while varying the window-size and use the benchmark test subjects defined in Table \ref{tab:dataset} as the test set. Figure \ref{fig:window_var} demonstrates the change in performance for different window-size and it can be concluded from the figure that the proposed model is less sensitive to changes in window-size than the other models in terms of performance. It is evident that datasets involving complex activities require relatively longer time-span for sliding window for capturing correct activity label.

\textbf{Feature Map for Sensor Modality Attention:} Sensor modality attention layer described in Section \ref{proposed:sensor-modality} has been utilized to determine the impact of sensors' placements on the classification task. In Figure \ref{fig:featuremap}, feature attention maps incorporate average attention on specific sensor modality listed in the x - axis over all activity window. The vertical axis indicates timestep within a specific window. If we consider the feature maps visualizing attention weights on sensor modality, it can be derived that while ironing, sensors placed at hand get greater attention weights which is visualized in Figure \ref{fig:sub-first}. On the other hand, during ascending stairs, hand accelerometer and ankle accelerometer obtain relatively higher attention weight in feature segmentation which is evident in Figure \ref{fig:sub-second}.  Moreover, the attention map in Figure \ref{fig:sub-third} exhibits that gyroscope placed at chest obtains higher attention weight than other sensor placements. The activity cycling involves simultaneous movement of different body parts which is captured through sensor modality attention and evident in the attention weight distribution in Figure \ref{fig:sub-fourth}. Vacuum cleaning activity in \ref{fig:sub-fifth} indicates that hand is the dominant body part in detecting this particular activity. The attention map illustrates similar weight distribution as ascending stairs in the case of walking. 

Figure \ref{fig:featuremap} demonstrates the higher emphasis on particular sensor modality in predicting class label. Here, the proposed model automatically distributes attention weight on heterogeneous sensor modalities and this weight is intuitively explainable with respect to the predicted class label. 

\section{Conclusion}
\label{conclusion}
In this paper we propose a self-attention-based deep learning architecture for Human Activity Recognition (HAR). The model is adapted from the transformer architecture proposed for machine translation. The proposed model foregoes recurrent layers and utilizes attention mechanisms to generate feature representation used for classification. We perform experiments on leave one subject out cross validation on four benchmark datasets - PAMAP2, Opportunity, Skoda, and USC-HAD. We perform both sample-wise and window-wise classification. Compared with existing state-of-the art methods, we show that our proposed attention based model outperforms existing models in case of the benchmark test subject for all datasets except Opportunity for sample wise classification. In case of window-wise classification our model outperforms Deep Convolutional LSTM and Convolutional Autoencoder models. One limitation of our experiments is that we did not perform window-wise classification on the newly published models. In future, we intend to extend our model with a decoder network and perform more extensive experiments to compare with all existing models.

\section*{Acknowledgements}
This work is supported by grants from ICT Division, Government of Bangladesh and Independent University, Bangladesh.

\bibliography{ecai}
\end{document}